\documentclass{article}

\usepackage[final]{nips_2018}

\usepackage[utf8]{inputenc} % allow utf-8 input
\usepackage[T1]{fontenc}    % use 8-bit T1 fonts
\usepackage{hyperref}       % hyperlinks
\usepackage{url}            % simple URL typesetting
\usepackage{booktabs}       % professional-quality tables
\usepackage{amsfonts}       % blackboard math symbols
\usepackage{nicefrac}       % compact symbols for 1/2, etc.
\usepackage{microtype}      % microtypography

\usepackage{amsmath}
\usepackage{amsthm}
\usepackage{amssymb}
\usepackage{graphicx}

\title{Fake News Detection with Different Models}

\author{%
Sairamvinay Vijayaraghavan\\
\texttt{saivijay@ucdavis.edu} \\
\And
Zhiyuan Guo \\
\texttt{zhyguo@ucdavis.edu } \\
\And
Ye Wang \\
\texttt{kyewang@ucdavis.edu} \\
\AND
John Voong \\
\texttt{jhvoong@ucdavis.edu } \\
\And
Wenda Xu \\
\texttt{wedxu@ucdavis.edu} \\
\AND
Armand Nasseri \\
\texttt{aanasseri@ucdavis.edu} \\
\And
Jiaru Cai \\
\texttt{jrcai@ucdavis.edu } \\
\AND
Linda Li  \\
\texttt{tsqli@ucdavis.edu } \\
\And
Kevin Vuong \\
\texttt{kkvuong@ucdavis.edu } \\
\And
Eshan Wadhwa \\
\texttt{ewadhwa@ucdavis.edu} \\
}

\begin{document}
\maketitle

\begin{abstract}
\textbf{Problem: }The problem we intend to solve is modelled as a binary classification problem. We intend to find the relation in the words and the context in which the words appear within the text and how it could be used to classify texts as real (negative cases) or fake (positive).

\textbf{High-level description: }Many news sources contain false information and are therefore “fake news.” Because there is a lot of “fake news” articles and fabricated, misleading information on the web, we would like to determine which texts are legitimate (real)  and which are illegitimate (fake). To solve this as a binary classification problem, we investigate the effectiveness of different Natural Language Processing models which are used to convert character based texts into numeric representations such as TFIDF, CountVectorizer and Word2Vec models and find out which model is able to preserve most of the contextual information about the text used in a fake news data set and how helpful and effective it is in detecting whether the text is a fake news or not.

\textbf{Results:}We find that out of the three pre-training vectorizing algorithms, Word2Vec performs comparatively the worst in general and the CountVectorizer performs slightly better than the TF-IDF models in most of the cases. Out of
the five fine-tuning algorithms, neural networks (ANNs and LSTMs) perform better. A combination of cv with LSTM achieves the best performance.

\textbf{Contribution to the machine learning field: }We presented a simple model which can be used to classify a given text as “real” or “fake” mostly accurately. This form of pre-training embedding algorithms and then fine-tuning on the downstream supervised task (of binary classification) proves to be efficient and effective in classifying susceptible news text.

\end{abstract}

\section{Introduction}
For this report, we are exploring the field of natural language processing, which is the broad study of how computers and machines can understand human to human communication and how texts are analyzed based on contextual information by machines.

In particular, we are using natural language processing to classify news articles as real news or “fake news”. Fake news is misinformation masked under the guise of a real news article, and is used to deceptively influence people’s beliefs. 

For this report, we are classifying news articles as “real”  or “fake”, which will be a binary classification problem - classifying the samples as a positive (with fake news) or negative (not fake news) sample. Many studies have used machine learning algorithms and build classifiers based on features like content, the author’s name and job-title, using lots of models like the convolutional neural network (CNN), recurrent neural network (RNN), feed-forward neural network (FFNN), long-short term memory (LSTM) and logistic regression to find the most optimal model and return its results. In [1], the author built a classifier using natural language processing and used models like CNN, RNN, FFNN, and Logistic Regression and concluded that the CNN classifiers could not be as competitive as the RNN classifiers. The authors in [2] think that their study can be improved by having more features like knowing the history of lies spoken by the news reporter or the speaker.

Moreover, apart from the traditional machine learning methods, new models have also been developed. One of the newer models, TraceMiner, creates an LSTM-RNN model inferring from the embedding of social media users in the social network structure to propagate through the path of messages and has provided high classification accuracy$^{5}$. FAKEDETECTOR is another inference model developed to detect the credibility of the fake news which is considered to be quite reliable and accurate$^{7}$. 

There also have been studies that have a different approach. A paper surveys the current state-of-the-art technologies that are imperative when adopting and developing fake news detection and provides a classification of several accurate assessment methods that analyze the text and detect anomalies$^{3}$.

These previous approaches lack a clear contextual analysis used in NLP. We considered the semantic meaning of each word and we feel that the presence of particular words influence the meaning. We reckoned this important since we felt the contextual meaning of the text needs to be preserved and analyzed for better classification. Other studies emphasize the user and features related to them. In [4], “45 features…[were used] for predicting accuracy...across four types: structural, user, content, and temporal,” so features included characteristics beyond the text. Article [6] "learn[s] the representations of news articles, creators and subjects simultaneously." In our project, we emphasize the content by working with articles whose labels only relate to the text, nothing outside that scope, and have used SVM, Logistic Regression, ANN, LSTM, and Random Forest.

We had devised this problem into 3 different phases: pre-processing, text-to-numeric representation conversion using pre-trained algorithms, and then evaluate the models using state-of-the-art machine learning algorithms. We had analysed the data set and in particular the text part of the data explaining how it is distributed and then we converted each text into numeric representation using pre-training models such as TFIDF, CV and W2V for vector representation. Finally, we evaluated our numeric conversion data using significant machine learning algorithms such as neural networks, classification algorithms etc to perform the classification.

\section{Methods}

\subsection{The Dataset}
The training data set has five features: \textit{ID, title, author, text}, and \textit{label}. The ID uniquely identifies the news article. The title and author are the title and author of the news article respectively. The text is the content of the article, and may be incomplete. The label indicates whether the article is reliable (real) or not (fake):
\begin{center}
label = $\begin{cases}
0 & \textrm{if reliable news} \\
1 & \textrm{if fake news}
\end{cases}$
\end{center}
The training data set contains 20800 odd number of samples.

The test data set does not have labels, so we do not use it. The test data set will be selected from the training data set randomly when we are evaluating our models.

In our project, since we hypothesized that the text and the words used within the text are key to distinguish between real and fake news samples, we decided to investigate only the text column.

\subsection{Data Pre-processing}
\subsubsection{Removed numbers}
Within the context of a news article title or text, numbers simply quantify claims and do not change the meaning of the text. Therefore it is best to remove all numbers to minimize noise in our data. We use the \texttt{string.digits} string constant in Python as well as the \texttt{translate} and \texttt{maketrans} methods from Python’s \texttt{string} module to convert all numerical digits to an empty string, effectively removing all digits. 

\subsubsection{Removed punctuation and special characters}
In addition of pre-processing the textual data, we removed all characters that are not textual (not alphabets such as punctuation, extra delimiters etc.). We used the \texttt{string.punctuation} module in Python to find all punctuation characters. We remove all those punctuation characters from every word in the texts, with the exception of the symbols ‘\#’ and ‘@’. Because these are characters used for Twitter hashtags and mentions, we handle these later. Next, we removed an assortment of special characters that don’t appear on traditional American keyboards and don’t contribute to the meaning of the tweets. The long dash (“–”), single and double Asian quotations, ellipse characters (…), and bullet points (•) all were removed for this reason. 
 
After removing all special characters, there are still a couple of pre-processing cases we account for. For these cases, we used regular expressions to detect certain patterns we wish to remove. One of the patterns is Twitter hashtags and mentions. In a news setting, Twitter hashtags and mentions are often added to try to obtain more search results and relevance, but often distract from the overall meaning of the news content itself. In our problem, we are primarily concerned with words and mostly their contextual meanings used in the text and we assumed that these unnecessary characters. To detect the hashtags and mentions, we simply use regular expressions to remove all text after a hashtag (\#) or @ symbol, and stop removing text when we reach the next space. We also use regular expressions to handle em dashes (—) and more than two consecutive spaces. Em dashes are used in various linguistic contexts like joining independent clauses. They do not add to the meaning of the text, however they are surrounded by two words of different clauses, so we replaced all em dashes with a single space to maintain the integrity of each phrase. Lastly, we replace any set of two or more consecutive spaces with just one space. 

Proceeding further, we make all of our texts lowercase and then remove all rows that have foreign language characters in their text, since we are only interested in identifying fake news in English. To do this we used the package \texttt{langid} in Python to identify the language of all texts, and removed all rows with foreign characters. This finally ensures the text we preserve is only with English words with no non-alpha character.

\subsubsection{Removed stop words}
Stop words are a list of the most common words in a language, such as “a”, “be”, “quite”, “should”...etc. They are often void of meaning, and does not add anything to the content. They are also most frequently present in every text. Hence, we presumed removal of stop words can have multiple advantages. For once, it decreases memory overhead, since we cut down a huge amount of text (and hence narrows down the number of features to train our models on). Second, it reduces noise, since by eliminating stop words, we are able to focus on more meaningful contents (the more distinct features between these two classes). Although it is not often the case that removing stop words are the most optimal, sometimes the information that we are looking for may be included in the stop words that we removed. For example, in most cases of language modeling, or translation, where it is important that we keep all the stop words. However, in our circumstances, we are using the semantics of the text to make a decision. In this case, we can safely remove stop words to observe the more meaningful context words.

\subsection{Data Distribution}
We performed some data analysis on the text and wanted to understand how the text is distributed. We had analyzed and represented our data (text) distribution in a few different perspectives. We first analyzed the data through graphing its sentiment polarity, most popular unigram and bigram, as well as looking at the distribution of the word types. We will be comparing the graphs before and after preprocessing, which includes, stop word removal, removing punctuation and special characters, and numbers. 

\subsubsection{Sentiment Polarity}
\begin{center}
\textit{Polarity Graphs before pre-processing}\\
\includegraphics[scale = 0.42]{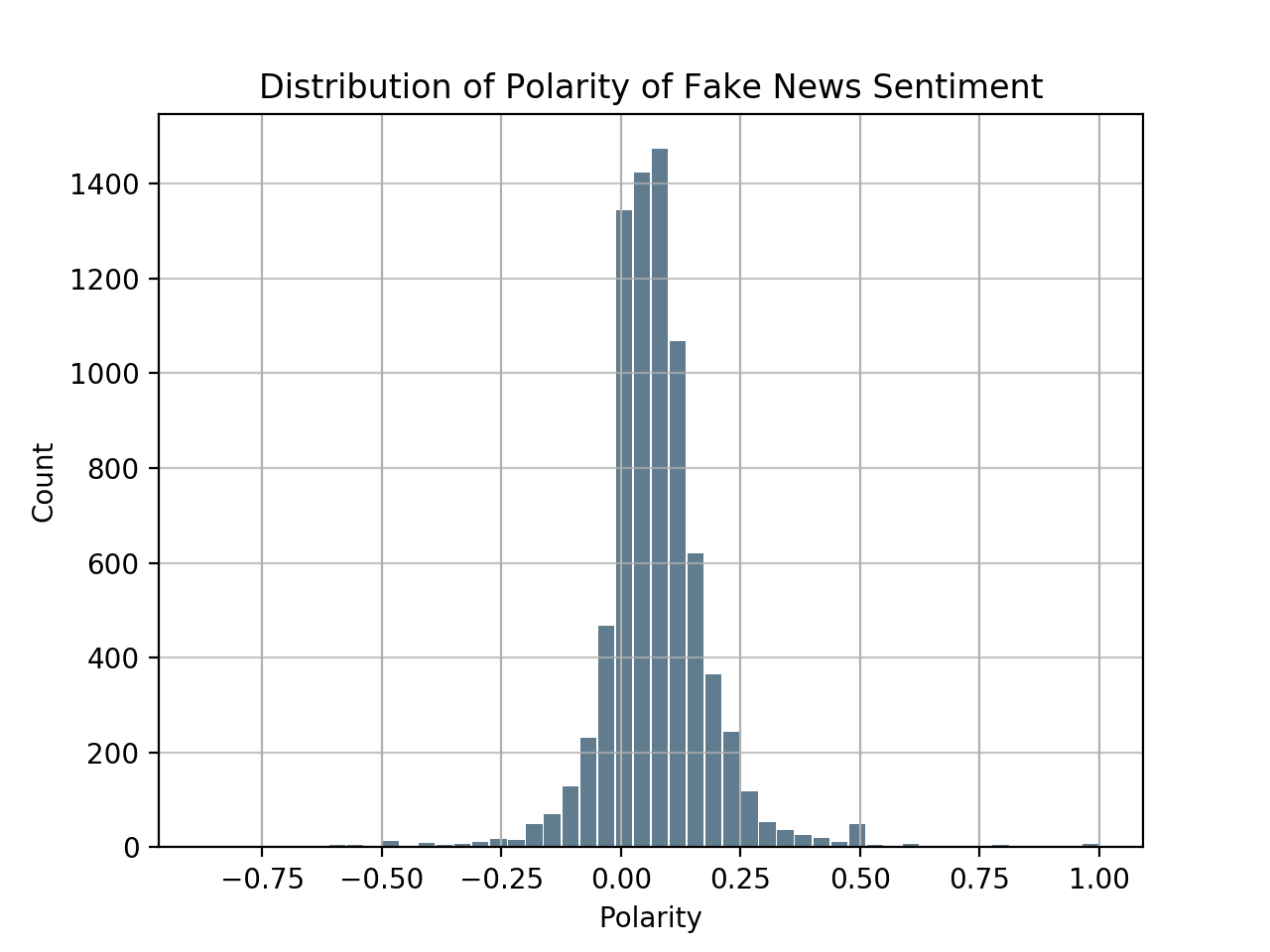}
\includegraphics[scale = 0.42]{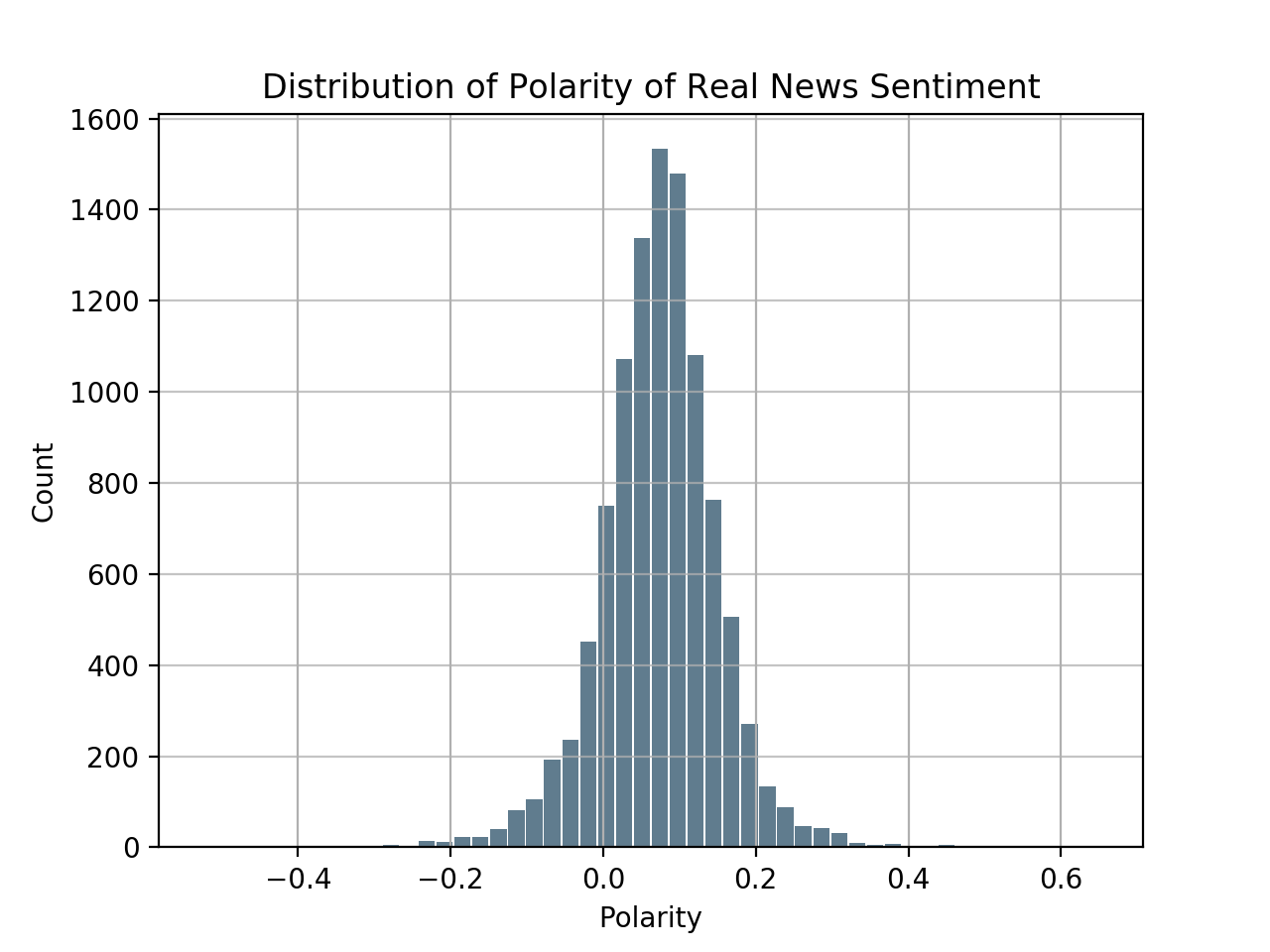}\\
\textit{Polarity Graphs after pre-processing}\\
\includegraphics[scale = 0.34]{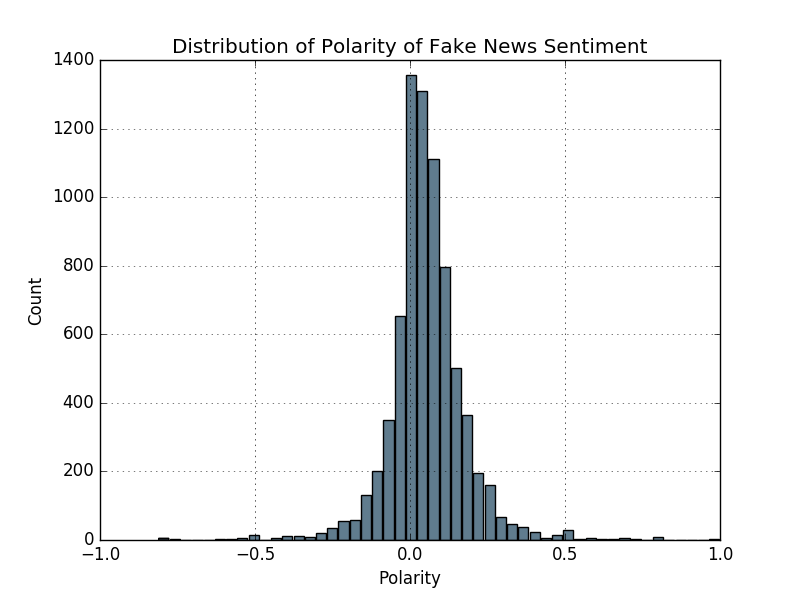}
\includegraphics[scale = 0.34]{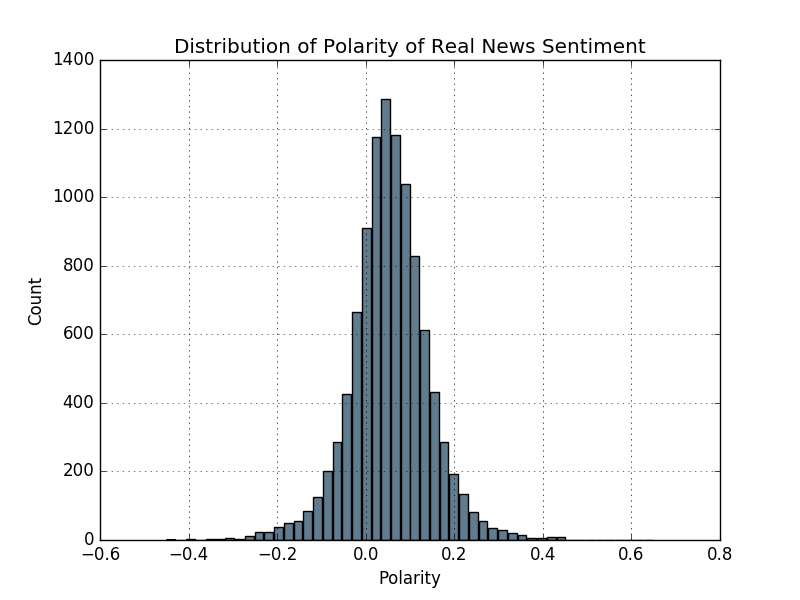}
\end{center}
For both before and after pre-processing, the distribution of the polarity of fake news sentiment and real news sentiment are mostly the same. For both fake news and real news, there are slightly more positive news than the negatives. However, there is a noticeable difference between the polarity. We can see that although not by much, fake news are a little bit more polar than real news. There are more outliers, and the data are a little bit more spread out.

\subsubsection{Part of Speech Distribution}
\begin{center}
\textit{Part of Speech Graphs before pre-processing}\\
\includegraphics[scale = 0.42]{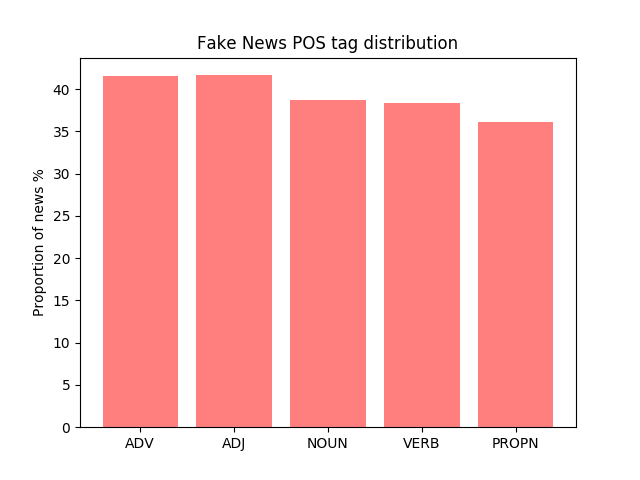}
\includegraphics[scale = 0.42]{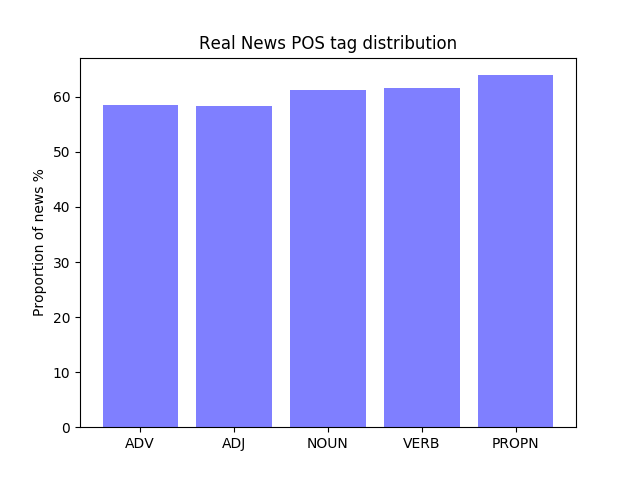}
\newpage
\textit{Part of Speech Graphs after pre-processing}\\
\includegraphics[scale = 0.34]{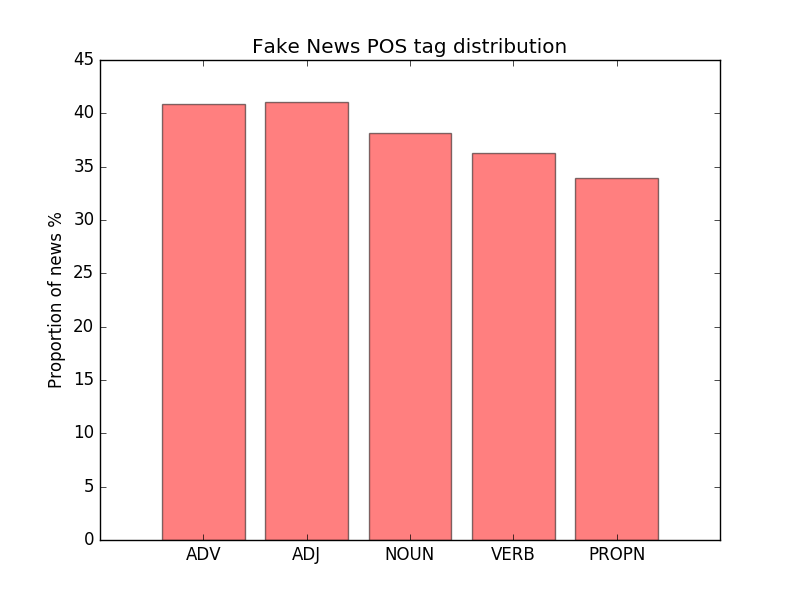}
\includegraphics[scale = 0.34]{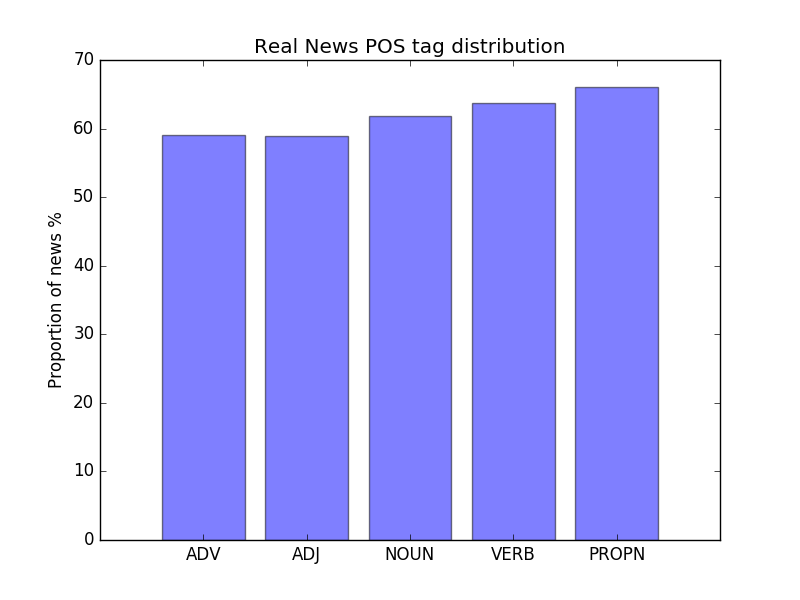}
\end{center}

Although the differences are slight, there is a difference in part of speech distribution between real and fake news. In fake news, there are a higher percentage of adverbs and adjectives compared to all the other parts of speech, while there is a lower percentage of proper pronoun; however, in real news, there are a higher percentage of pronoun. We can interpret this as there are more adverbs and adjectives in fakes new, and there are more pronoun in real news. Perhaps, this is indicating that fake news are more likely to use adverbs and adjectives to embellish their sentences, while real news use more pronouns to establish as reference to their legitimacy.

\subsubsection{Unigram and Bigram}

\begin{center}
\large \textit{Unigrams}\\
\normalsize
\begin{tabular}{cccc}
\toprule
\multicolumn{2}{c}{Real News}   & \multicolumn{2}{c}{Fake News}             \\ \cmidrule(r){1-2} \cmidrule(r){3-4} 
Before   & After           & Before          & After              \\
\midrule
the       & nt           & the       & nt           \\
to        & trump        & to        & Trump        \\
of        & people       & of        & people       \\
and       & clinton      & and       & clinton      \\
in        & hillary      & in        & hillary      \\
that      & said         & that      & said         \\
for       & like         & is        & like         \\
on        & new          & for       & new          \\
he        & time         & it        & time         \\
is        & World        & on        & world        \\
it        & state        & as        & state        \\
was       & election     & with      & election     \\
said      & government   & are       & government   \\
mr        & president    & this      & preseident   \\
with      & war          & by        & war          \\
as        & years        & before    & years        \\
his       & states       & was       & states       \\
at        & american     & you       & american     \\
by        & obama        & have      & obama        \\
from      & media        & they      & media        \\
\bottomrule
\end{tabular}
\end{center}

\newpage
\begin{center}
\large \textit{Bigrams}\\
\normalsize
\begin{tabular}{cccc}
\toprule
\multicolumn{2}{c}{Real News}   & \multicolumn{2}{c}{Fake News}             \\ \cmidrule(r){1-2} \cmidrule(r){3-4} 
Before   & After           & Before          & After              \\
\midrule
of the   & mr trump        & of the          & hillary clinton    \\
in the   & united states   & in the          & donald trump       \\
to the   & new york        & to the          & united states      \\
on the   & mr trumps       & on the          & white house        \\
mr trump & white house     & and the         & new york           \\
at the   & donald trump    & that the        & hillary clintons   \\
and the  & mrs clinton     & to be           & clinton campaign   \\
that the & said mr         & for the         & clinton foundation \\
to be    & york times      & it is           & secretary state    \\
he said  & islamic state   & with the        & nt know            \\
with the & mr obama        & from the        & american people    \\
from the & breitbart news  & by the          & mainstream media   \\
by the   & president trump & at the          & foreign policy     \\
it was   & years ago       & hillary clinton & bill clinton       \\
\bottomrule
\end{tabular}
\end{center}
The comparison between the result of the top unigram and bigram before and after preprocessing demonstrates that our decision to remove stop words is the correct choice. The top unigram and bigram are all consisted of  words, in other words, filler words that does supply us with any explanation.

After removing the stop words, we can see that the top unigrams and bigrams become much more specific. 

\subsection{Unsupervised Pre-training to encode our texts into numeric representations}
\begin{center}
\includegraphics[scale = 0.4]{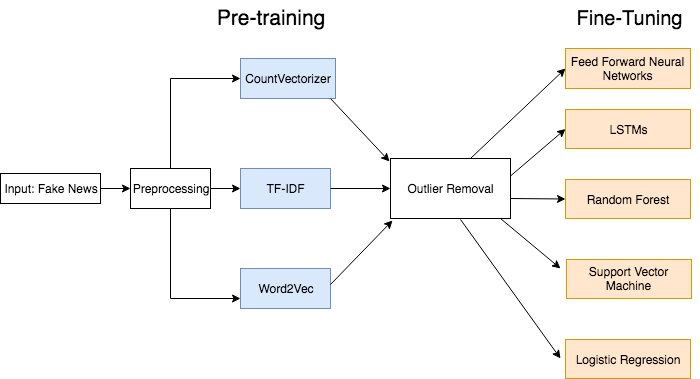}
\end{center}
\subsubsection{Natural Language Processing Models}
After text have been cleaned, they are mapped into numeric representations in form of vectors of the textual data using three pre-training algorithms (i.e. \textit{CountVectorizer}, \textit{TF-IDFVectorizer}, and \textit{Word2Vec}). Each sample, originally consisting of all text, is converted into a vector of features. Since only the text is passed into these pre-training algorithm, this stage is unsupervised. In the cases of CountVectorizer and TfidfVectorizer, the number of features is clipped at 10000 to avoid memory overrun and overfitting (because of the large number of features (the vocabulary)).

\subsubsection{CountVectorizer}
The CountVectorizer provides a simple way to both tokenize a collection of text documents and build a vocabulary of known distinct words, but also to encode new documents using that vocabulary$^{13}$.

Given a collection of text documents, $S$ , CountVectorizer will generate a sparse matrix $A$ of size $m$ by $n$, where $m =$ total number of documents, $n =$ total number of distinct words used in $S$.
\begin{center}
$A = \begin{pmatrix}
a_{11} & a_{12} & \cdots & a_{1n}\\
\vdots & \vdots & \vdots & \vdots \\
a_{m1} & a_{m2} & \cdots & a_{mn}
\end{pmatrix}$
\end{center}
This matrix is the one hot encoded representation of the different words present in the corpus. Entry $a_{ij} =$ total number of times $j$th word appears in the $i$th document.

We had converted the sparse matrix into a dense one since we found that there are plenty of distinct words in the corpus which may not even be present in some of the samples and hence they may be populated with zeros. Hence, we felt that since zeros may be entirely populated, we decided to convert it to a dense matrix using the \texttt{todense()} method call which a dense representation of the sparse matrix.

\subsubsection{TF-IDFVectorizer}
Although TF-IDF is an old algorithm, it is simple and effective to be used in the phase of pre-training$^{11}$. The computation of TfidfVectorizer involves computing the product of term frequency and inverse document frequency.  As the term implies, TF-IDF calculates values for each word in a document through an inverse proportion of the frequency of the word in a particular document to the percentage of documents the word appears in$^{12}$. 

The term frequency $tf(t, d)$ calculates the proportion of times that the term $t\in V(d)$ appears in the document $d$. The vocabulary $V(d) = \sum_t n(t,d)$ is constructed by the document $d$. Thus, if a word $w'$ does not appear in a document $d'$, the term frequency $tf(t', d')$ in this case would be zero. The idea of the term frequency is essentially the same as CountVectorizer.

$$tf(t,d) = \frac{n(t,d)}{V(d)}$$
$$n(t,d) = \textrm{ occurrence of the word }t\textrm{ in the document }d$$

Given a document collection $D$, the inverse document frequency $idf(t, D)$ is the log of the number of documents $N$ divided by $df(t,D)$, the number of documents $d \in D$ containing the term $t$. As a result, common words in $D$ will have a low term frequency score, while infrequent words will have a high term frequency. Thus, the term frequency will be very likely to separate fake news that often have less common words (even ungrammatical) from real news that usually consist of common words.
$$idf(t,D) =\log \Big(\frac{N}{df(t,D)}\Big)$$

As a summary, TF-IDF score $w(t,d)$ for a word increases with its count, but will be counteracted if the word appears in too many documents. 
$$w(t,d) = tf(t,d) \times idf(t, D)$$

Similar to CountVectorizer, we found that most of the entries within the matrix were 0. Hence, we used the dense (\texttt{todense()} call) to return the dense representation of the sparse TFIDF matrix representation.

\subsubsection{Word2Vec}
Word2Vec is another state of the art model used to represent words into vectors. Word2Vec is a simple neural network which basically tries to predict the next word within a context given a set of words provided. Word2Vec basically represents a vector for each word within the context and the vector representation is the weights of the particular connection from the input layer node into one of the hidden layer neurons. This information is mainly encoding the contextual information of the particular word within the corpus (collection of texts) on which we train our word2vec model.

In this project, all we did was we trained the word2vec model on our current corpus. We did this because we felt that the corpus contained very specific words which had a contextual meaning completely different from what is used in general. Hence, we chose to train the corpus on the existing texts in our corpus texts over the pre-trained word2vec models such as google models. For training our word2vec models, we chose the minimum count as the average number of words in each of the texts in general, since we believed that texts which are shorter than the mean length have less context and hence we rejected those sentences to train on. We then used the number of features as the default number of features as 100 since we wanted to analyze on a short number of features.

For this project, we decided on a very simple and plain approach. We obtained the vector for each sentence by summing all the vector representations for each word in the sentence only if the word belongs to the word2vec model. The summed up vector is finally divided with the number of words in the sentence since we wanted to make sure that the size of the text doesn’t affect the vector embeddings and hence we normalized our word2vec embedding.

\subsection{Outlier Removal}
During outlier removal, the Isolation Forest algorithm isolates observations by randomly selecting a feature and then randomly selecting a split value between the maximum and minimum values of selected features. In Isolation Forest, an anomaly score can be calculated as the number of conditions required to separate given observation. 
\begin{center}
\includegraphics[scale = 0.5]{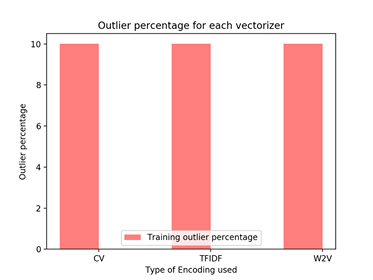}
\end{center}
In our outlier detections and removals, Isolation Forest has been applied to three different features. Generated from TFIDF, CV, WV. Percentages of outlier in each feature set is calculated, bar graph of percentage of training outliers are included.

\subsection{Fine-tuning}
Once the representations of text are pre-trained from previous unsupervised learning, the representations are then fed into 5 different models to perform supervised learning on the downstream task. In this case, the downstream task is a binary classification of the fake news as either real or fake. A k-fold prediction error is obtained from each of the 5 models, and since we have 3 different pre-training models, we have a total of 15 models to compare.

\subsubsection{Artificial Neural Network (ANN)}
We trained simple Artificial Neural Networks which contains an input layer, particular number of output layers (specified by a hyperparameter) in which each hidden layer contains the same number of neurons and the same activation function, and an output layer with just one node for the classification (real or fake) which uses \textit{sigmoid} as an activation function. We chose sigmoid as the output layer activation and the \textit{binary\_crossentropy} as the loss since it is a binary classification problem and the use of \textit{softmax} normalizes the results which is not needed for this problem and since we use only one output node to return the activation, we applied sigmoid for the output layer activation. We performed Grid Search strategy to find the best hyper-parameters such as activations, optimizers, number of hidden layers and number of hidden neurons. We had used \texttt{Keras Sequential} model and we used \texttt{Dense Layers} which contains connections to every hidden node in the next layer. 

Due to the limitation of computing resource, the grid search for Neural Networks is divided into three sequential steps. Instead of performing grid search on all the hyperparameters all at once, we chose to do grid search for the activations for the hidden layers, optimizers and the number of hidden layers and hidden neurons (done together). We coupled the number of hidden layers and the number of neurons since we believed that each of these hyperparameters interact with each other in improving the model training. We also did a \textit{K-fold} Split for 3 splits at each step and picked the best hyperparameters which renders the highest accuracy.

\subsubsection{Long Short Term Memory networks (LSTMs) }
Long Short Term Memory networks (LSTMs) is a special recurrent neural network (RNN) introduced by Hochreiter \& Schmidhuber (1997)$^{8}$.
\begin{center}
\includegraphics[scale = 0.35]{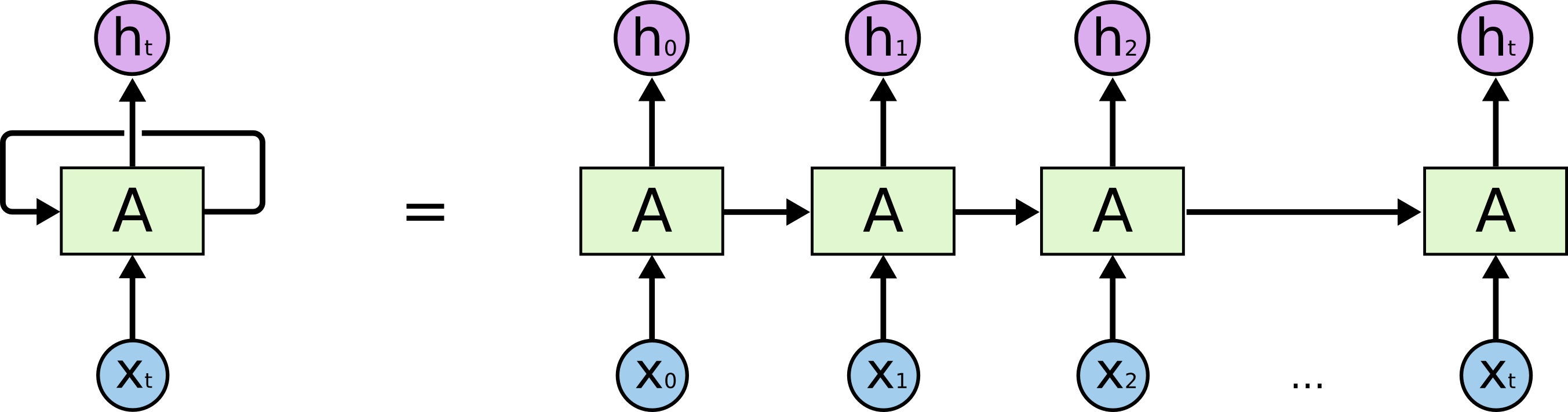}\\
\small
(Christopher Olah. “Understanding LSTM Networks.”)
\end{center}
The chain-like nature of an RNN allows information to be passed from the beginning all the way to the end. The prediction at time step $t$ depends on all previous predictions at time step $t’ < t$. However, when a typical RNN is used in a larger context (i.e. a relatively large time steps), the RNN suffers from the issue of vanishing gradient descent $^{9}$. LSTMs, a special kind of RNN, can solve this long-term dependency problem. 
\begin{center}
\includegraphics[scale = 0.35]{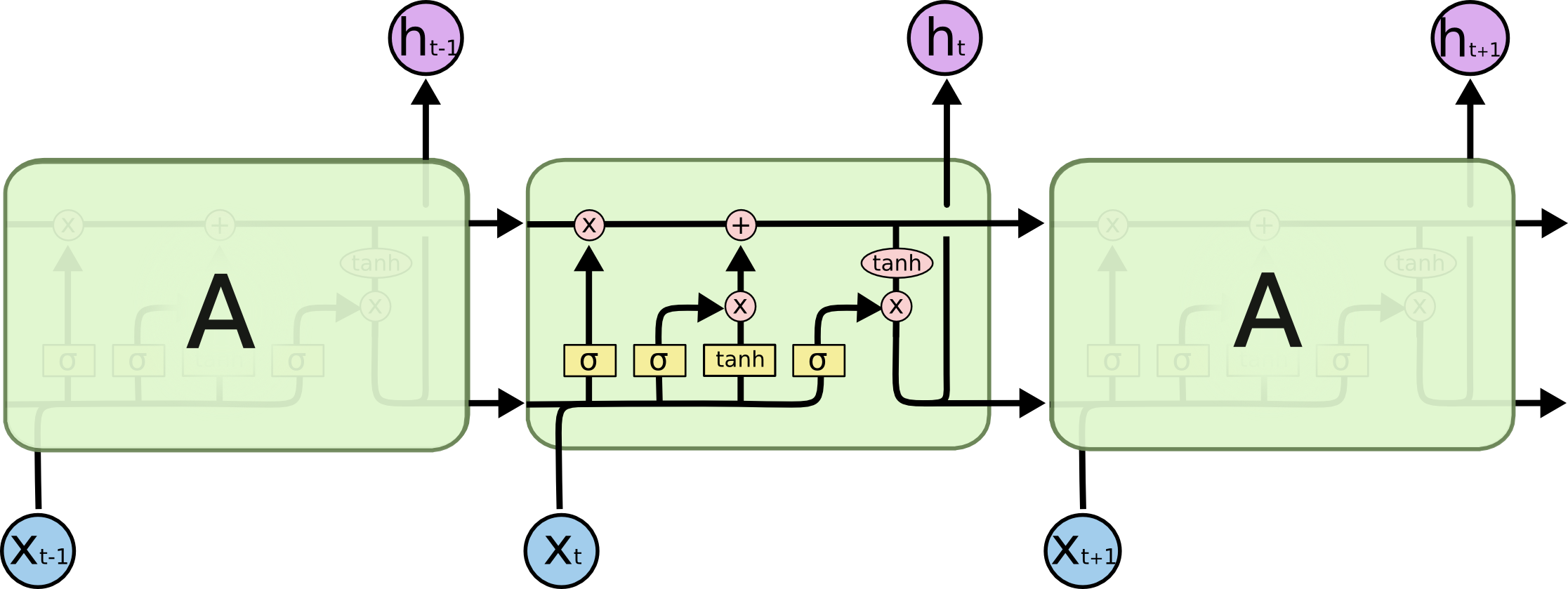}\\
\small
(Christopher Olah. “Understanding LSTM Networks.”)
\end{center}
Each cell in a typical LSTMs network contains 3 gates (i.e., forget gate, input gate, and output gate) to decide whether or not information should be maintained in the cell state  $C_t$.

For CountVectorizer and TfidfVectorizer, each sample of text is converted into a 1-d feature vector of size 10000. As a result, the number of time steps (i.e. the maximum amount of word vectors for each sample) for these two can only be set to 1, as the pre-trained representations are done at the sample’s level. By contrast, the number of time steps for Word2Vec can either be 1, if we simply take an average of the word embeddings, or the length of the sentence, where each word has an embedding and thus the pre-trained representations are done at the word’s level. We choose the approach with 1 timestep in our model because it requires less computation power. Meanwhile, we also do the length of the sentence, and 200 time steps are chosen as 200 is close to the mean amount of words in each sample and it is a fairly common choice in practice. However, since we do not have enough computation power to fine-tune (grid search) our model, we leave it out for our model and include it only in the final section.

In the LSTM layer, a dropout rate of 0.2, a common choice in practice$^{10}$ , is used to prevent overfitting. Grid search is performed in order to pick decent values of hyperparameters, including the number of hidden units in the LSTM layer, the number of hidden layers, the activation functions and the number of nodes in the hidden layer, and the optimizer. Relatively small numbers of hidden layers (i.e., \{0, 1, 2\}) and nodes (i.e., \{200, 400, 600\}) are selected as the basis for grid search, because this is a simple binary classification task and too many of them would cause overfitting.

Due to the limitation of computing resource, the grid search for LSTMs is divided into four sequential steps. Instead of performing grid search on all the hyperparameters all at once, the grid search is first done on the number of hidden layers and all other hyperparameters are randomly selected from the subset. Then, the grid search is done on the number of nodes in the hidden layer(s), using the best number of hidden layer found in step 1. The grid search completes when all four steps are finished. In each step we used \textit{K-fold} cross validation with $K = 3$.

\subsubsection{Random Forest}
A random forest is an ensemble classifier that estimates based on the combination of different decision trees. So random forest will fit a number of decision tree classifiers on various subsamples of the dataset. A random best subsets are built by each tree in the forest.  In the end, it gives the best subset of features among all the random subsets of features. 

In our project, 3 random forest algorithms have been applied with models count vectorizer, tfidf and word-to-vector. Random forest algorithm requires 4 hyperparameters to tune, such as the number of trees in the forest (i.e., \{200, 400, 800\}); the maximum depth of the tree (i.e., \{1,5,9\});  the minimum number of samples required to be at a lead node (i.e., \{2, 4\}); The minimum number of samples at each leaf node has the effect of smoothing the model, especially during regression; the minimum number of samples required to be at a leaf node (i.e., \{5, 10\}). All parameters are applied to grid search and in the end, the best set of parameters can be determined as we used \textit{K-fold} cross validation with $K = 3$.

\subsubsection{Logistic Regression}
Logistic regression is a statistical machine learning algorithm that classifies the data by considering outcome variables on extreme ends and this algorithm is providing a discriminatory line between classes. Compared to another simple model, linear regression, which requires hard threshold in classification, logistic regression can overcome threshold values for a large dataset. Logistic regression produces a logistic curve, which is limited to values between 0 to 1, by adding sigmoid function in the end. 

In regards to our project, three logistic regressions have been applied with models CountVectorizer, TF-IDF and Word2Vec. We did grid search on the solvers, including \textit{newton-cg, sag, lbfgs and liblinear}. Grid search is also performed on the inverse of regularization parameter with values being \{0, 4, 10\}. Best parameter sets can be determined as we used \textit{K-fold} cross validation with $K = 3$.

\subsubsection{Support Vector Machine (SVM)}
SVM is a supervised machine learning algorithm in which a hyperplane is created in order to separate and categorize features. The optimal hyperplane is usually calculated by creating support vectors on both sides of the hyperplane in which each vector must maximize the distance between each other. In other words, the larger the distance between each vector around the hyperplane, the more accurate the decision boundary will be between the categories of features.

In regards to our project, we fit 3 support vector machines on CountVectorizer, TfidfVectorizer, and WordToVectorizer. An SVM requires specific parameters such as a kernel type, $C$, maximum iterations, etc. In our case, we needed to determine the optimal $C$ as well as the optimal kernel for each fit. We used \textit{K-fold} cross validation with $K = 3$. A grid search of kernel types and $C$ was performed in order to give us the most accurate svm model. The parameters we used for each kernel were \textit{linear} and \textit{rbf} while the values we used for $C$ were 0.25 ,0.5, and 0.75. Once the grid search was completed for these hyperparameters, the model was evaluated with the most optimal hyperparameters using cross validation of 3 splits.

\section{Results}

\begin{center}
\large \textit{Grid Search Results}
\normalsize
\begin{tabular}{llll}
\toprule
 & CountVectorizer & TF-IDF & Word2Vec \\
 \cmidrule(r){2-4}
SVM & Kernel = Linear & Kernel = Linear & Kernel = Linear \\
 & C = 0.25 & C = 0.75 & C = 0.75 \\
  \midrule
Logistic Regression & Solver = sag & Solver = sag & Solver = newton-cg \\
 & C = 21.54 & C = 7.74 & C = 3593.81 \\
  \midrule
Random Forest & Max Depth = 9 & Max Depth = 9 & Max Depth = 9 \\
 & Min\_samples\_leaf = 2 & Min\_samples\_leaf = 4 & Min\_samples\_leaf = 2 \\
 & Min\_samples\_split = 10 & Min\_samples\_split = 5 & Min\_samples\_split = 10 \\
 & N\_estimators = 200 & N\_estimators = 400 & N\_estimators = 400 \\
  \midrule
ANN & Activation = relu & Activation = sigmoid & Activation = relu \\
 & Optimizer = Adam & Optimizer = Adam & Optimizer = Adam \\
 & Hidden\_layers = 2 & Hidden\_layers = 3 & Hidden\_layers = 1 \\
 & Num\_Neurons = 600 & Num\_Neurons = 400 & Num\_Neurons = 600 \\
  \midrule
LSTM & Activation = sigmoid & Activation = sigmoid & Activation = relu \\
 & Optimizer = Adam & Optimizer = Adam & Optimizer = Adam \\
 & Hidden\_layers = 2 & Hidden\_layers = 2 & Hidden\_layers = 2 \\
 & Memcells = 200 & Memcells = 200 & Memcells = 200 \\
 & Num\_Neurons = 200 & Num\_Neurons = 600 & Num\_Neurons = 600 \\
 \bottomrule
\end{tabular}

\large \textit{Mean Test Scores}
\normalsize
\begin{tabular}{|l|l|l|l|l|l|}
\hline
 & SVM & ANNs & LSTMs & LOGISTIC & RANDOM FOREST \\ \hline
CV & 93.06\% & 94.29\% &\textbf{ 94.88\%} & 94.45\% & 87.64\% \\ \hline
TFIDF & 94.58\% & 93.73\% & 93.89\% & \textbf{94.79\%} & 87.64\% \\ \hline
Word2Vec & 91.17\% & \textbf{93.06\%} & 92.29\% & 91.30\% & 88.60\% \\ \hline
\end{tabular}
\end{center}

\begin{center}
\includegraphics[scale = 0.12]{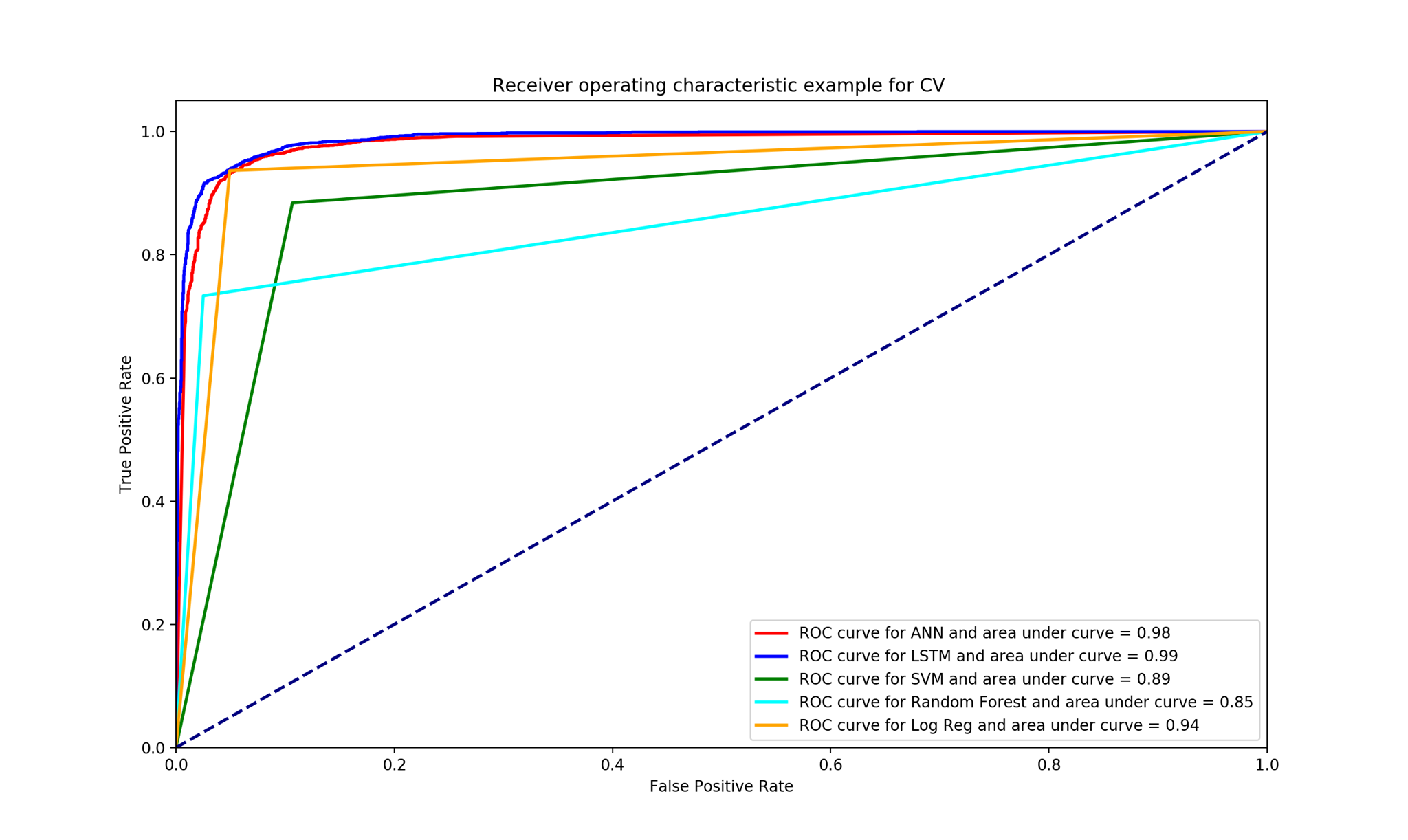}\\
\includegraphics[scale = 0.12]{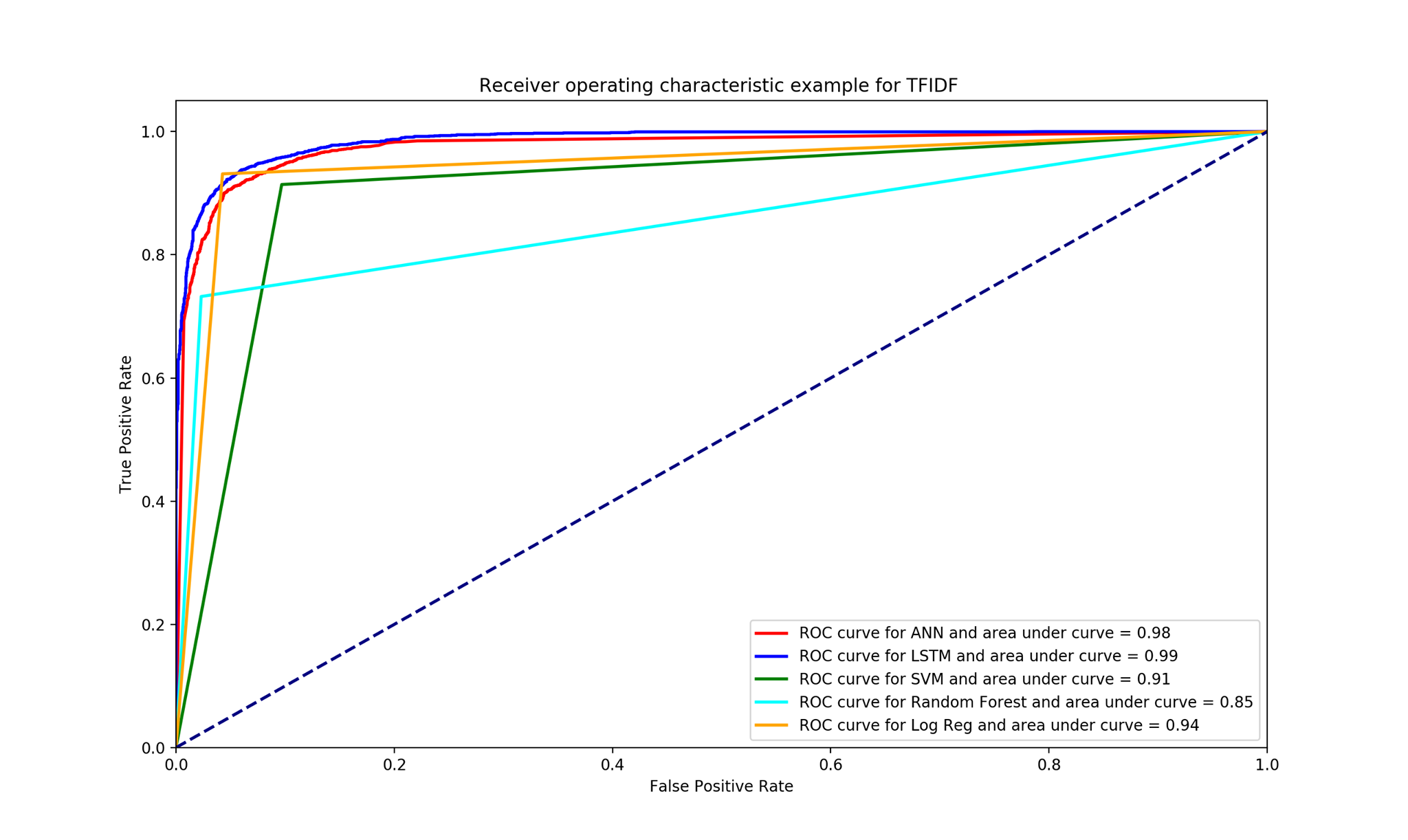}\\
\includegraphics[scale = 0.12]{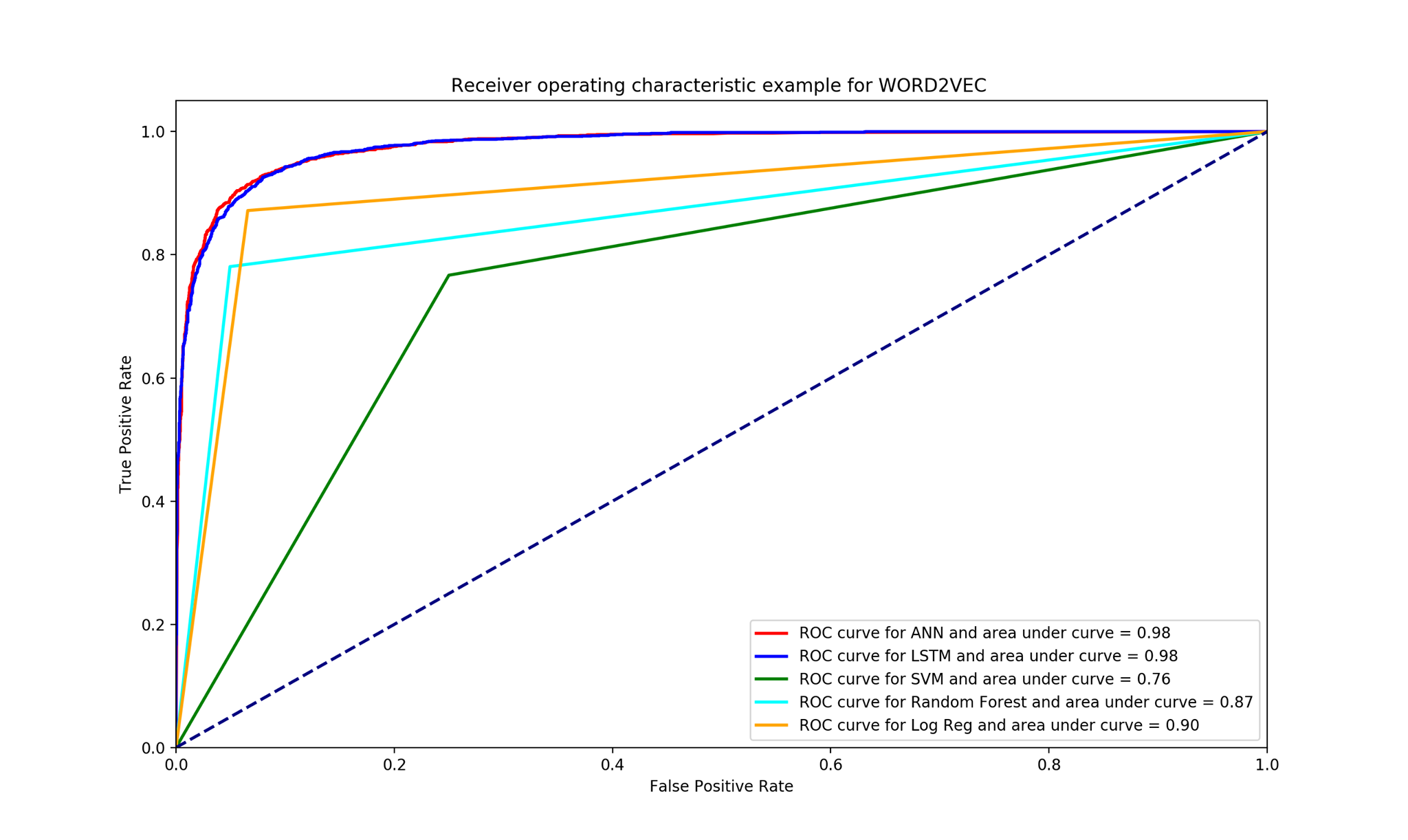}
\end{center}

\begin{center}
\textit{ANN Loss and Accuracy}\\
\includegraphics[scale = 0.42]{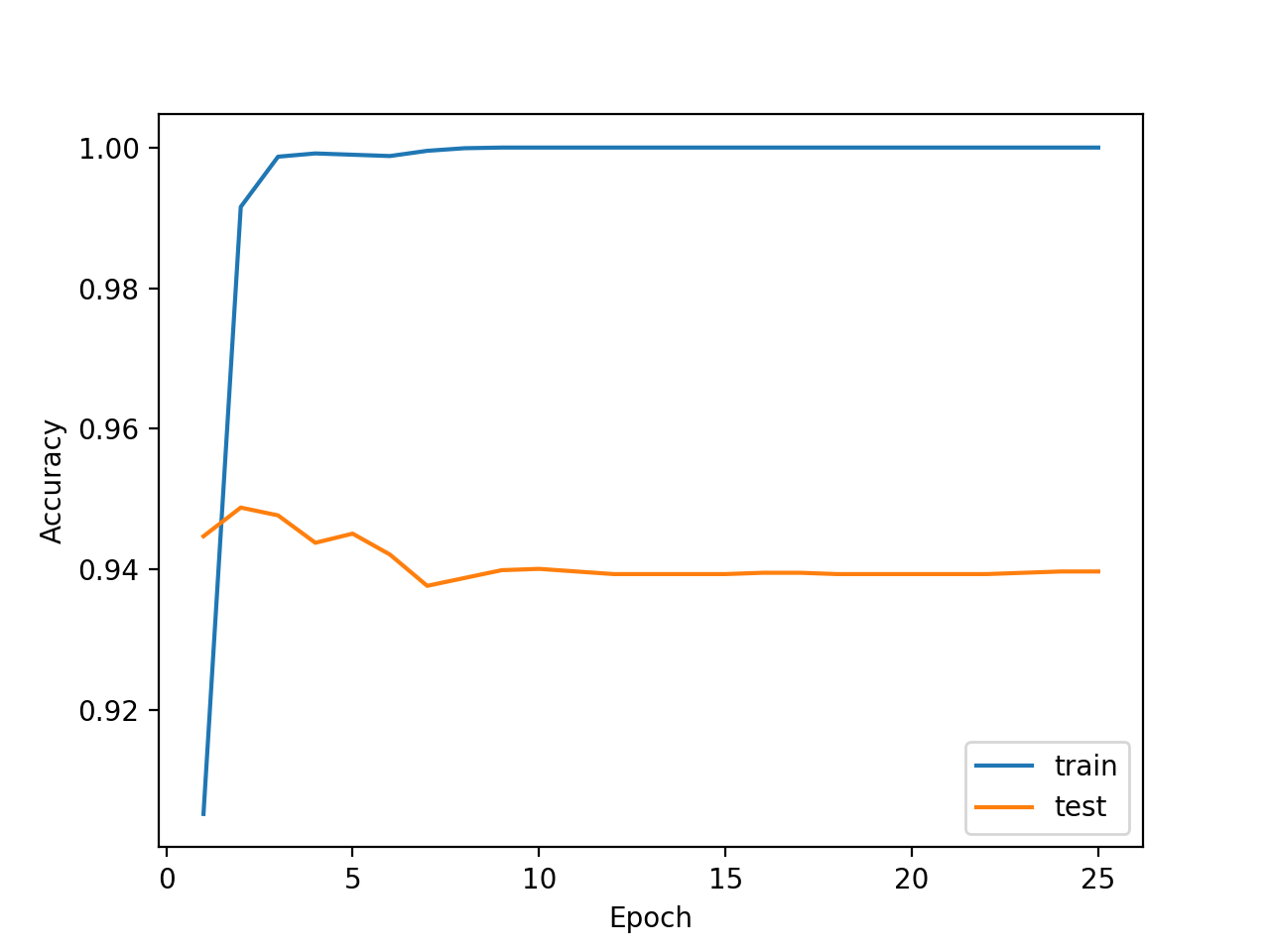}
\includegraphics[scale = 0.42]{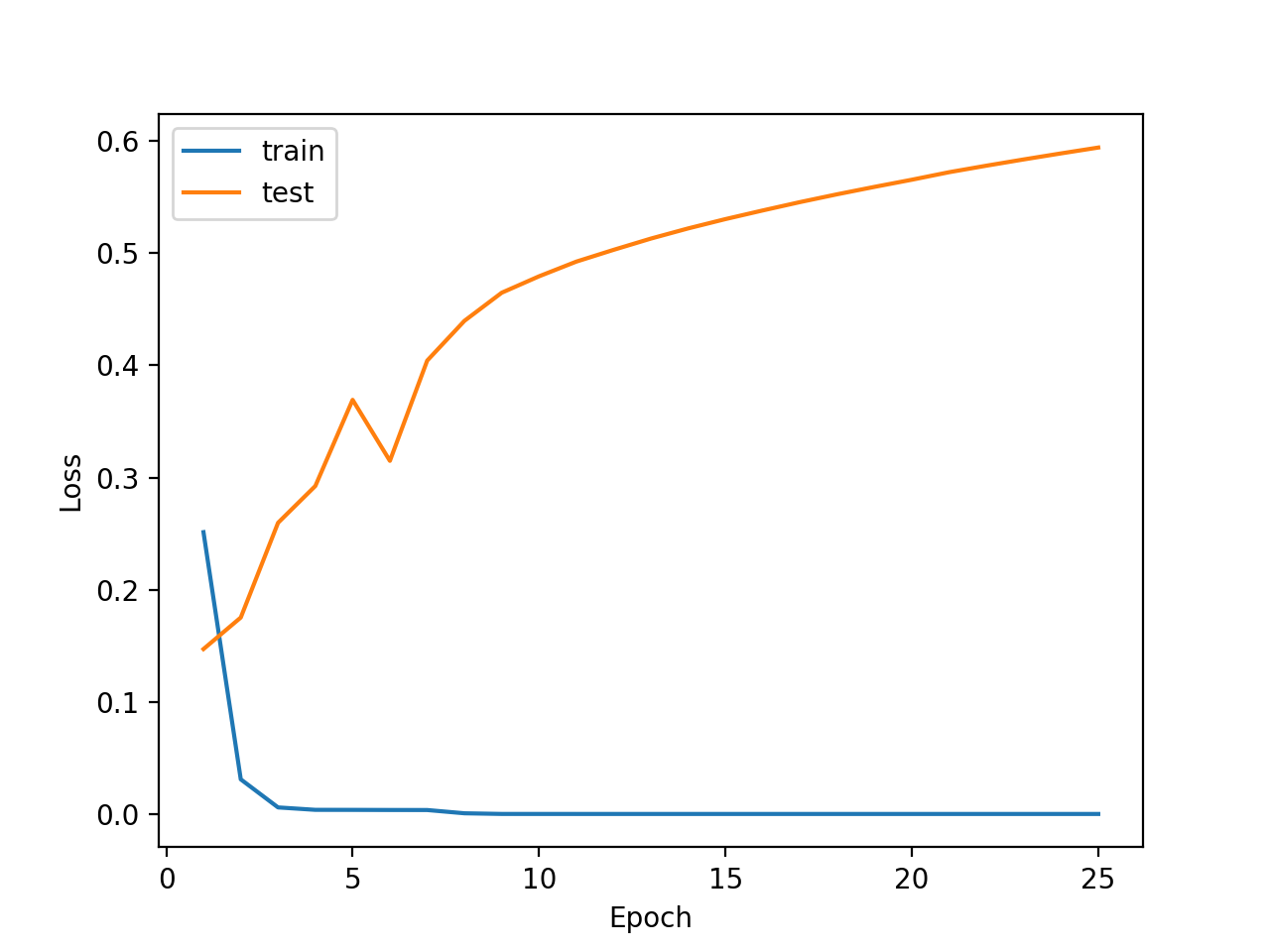}\\
\textit{LSTM Loss and Accuracy}\\
\includegraphics[scale = 0.42]{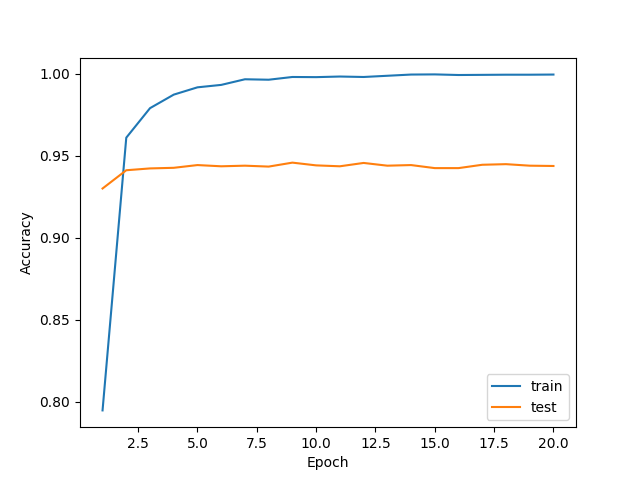}
\includegraphics[scale = 0.42]{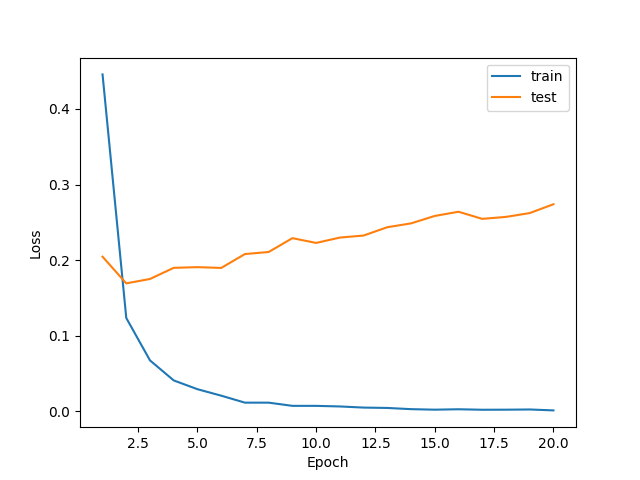}
\end{center}

The model is evaluated using a 3-fold of cross validation. Out of the fifteen models, CountVectorizer with LSTMs performs the best.
Word2Vec performs the worst among the three pre-training 
algorithms. Random forest performs the worst among the five 
fine-tuning algorithms.

\section{Discussion}
Among our three pre-training models, CountVectorizer achieves in general the best performance comparatively and Word2Vec performs relatively poor amongst the three models. The essential idea behind both CountVectorizer and TF-IDF is computing a score which depends on the frequency of the word belonging to the vocabulary. However, comparing to CountVectorizer, the TF-IDF includes an extra inverse document frequency that “penalizes” (apparently masks) the contextual meaning within the words that appear more frequently across documents. They represent the importance of the word within a document. The results may imply that even though the penalization is smoothed by a log function, the punishment may be too high.

The results also show that in general neural networks do the best consistently, as neural networks serve as a powerful universal approximator. However, the loss and accuracy plots show that we are using too many epochs and thus have the issue of overfitting. This is because our pre-training model is already very strong so it learns a good contextual representation of text. As a result, the epochs needed for downstream task are not much. In addition, one thing to note is that logistic regression also performs very well. This implies that our data are mostly linearly  separable. While neural networks can fit the data very well, but they run the risk of overfitting the data. As a result, neural networks are not as good as SVM and Logistic Regression for TF-IDF.

A combination of CountVectorizer and LSTMs is the best among all the models. While LSTMs with one timestep are very similar to ANN in terms of architecture, LSTMs have gates and a tanh activation function inside the module. This different design may let LSTMs perform slightly better than ANN.

Word2Vec does not perform well. One reason is that we are simply taking an average of the word embedding vectors to get a generalized vector representation of each sample of paragraph. Taking an average fails to represent the dependencies between words. Another reason is that we do not use pre-trained Word2Vec embeddings available online from huge corpus but instead build our own from the dataset. While we thought that building our own Word2Vec would make the model specific to this task, the results show that Word2Vec may need to be built from larger dataset.

\section{Conclusion}
This report provides a fairly simple approach to encode texts and how the presence of words in general impacts the classification of texts as real and fake.

We achieved high accuracy results in most of our algorithms and in particular neural networks generally do better than the others.

What's worth noting is that our LSTMs only use a timestep of 1 
and are essentially multi-layer perceptrons. Still, as mentioned 
is the LSTM's method section, the LSTMs with the real recurrence 
are performed by using Word2Vec for representations at the word's level. In this case, each word has its own vector, and a sample
will be a collection of vectors and thus a 2-D matrix. As mentioned before, each vectorized word will become a timestep, and a total of 200 timesteps is used (If the paragraph has more than 200 words, only the first 200 words will be selected).
We run our model and get the following results.

\includegraphics[scale=0.42]{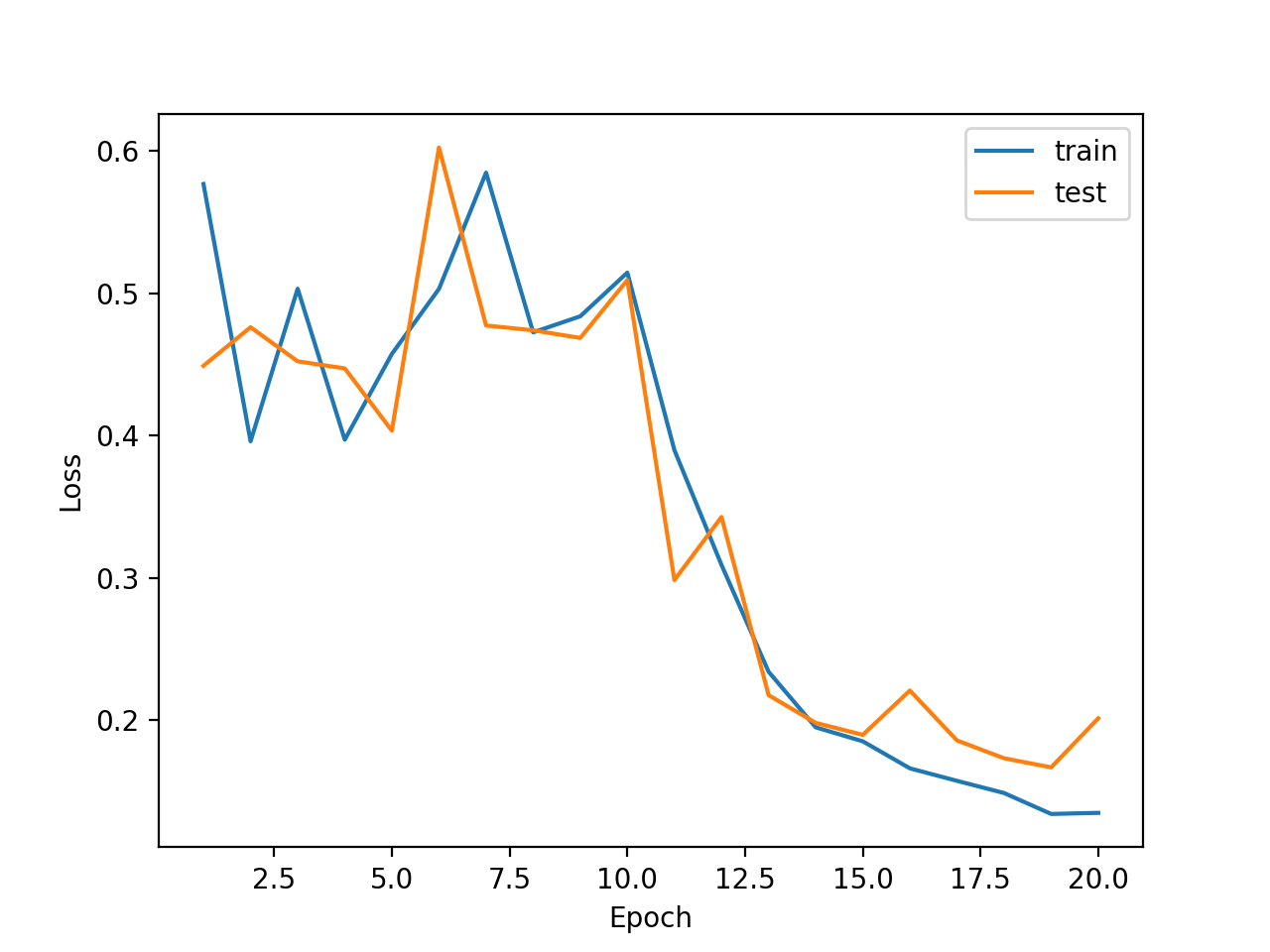}
\includegraphics[scale=0.42]{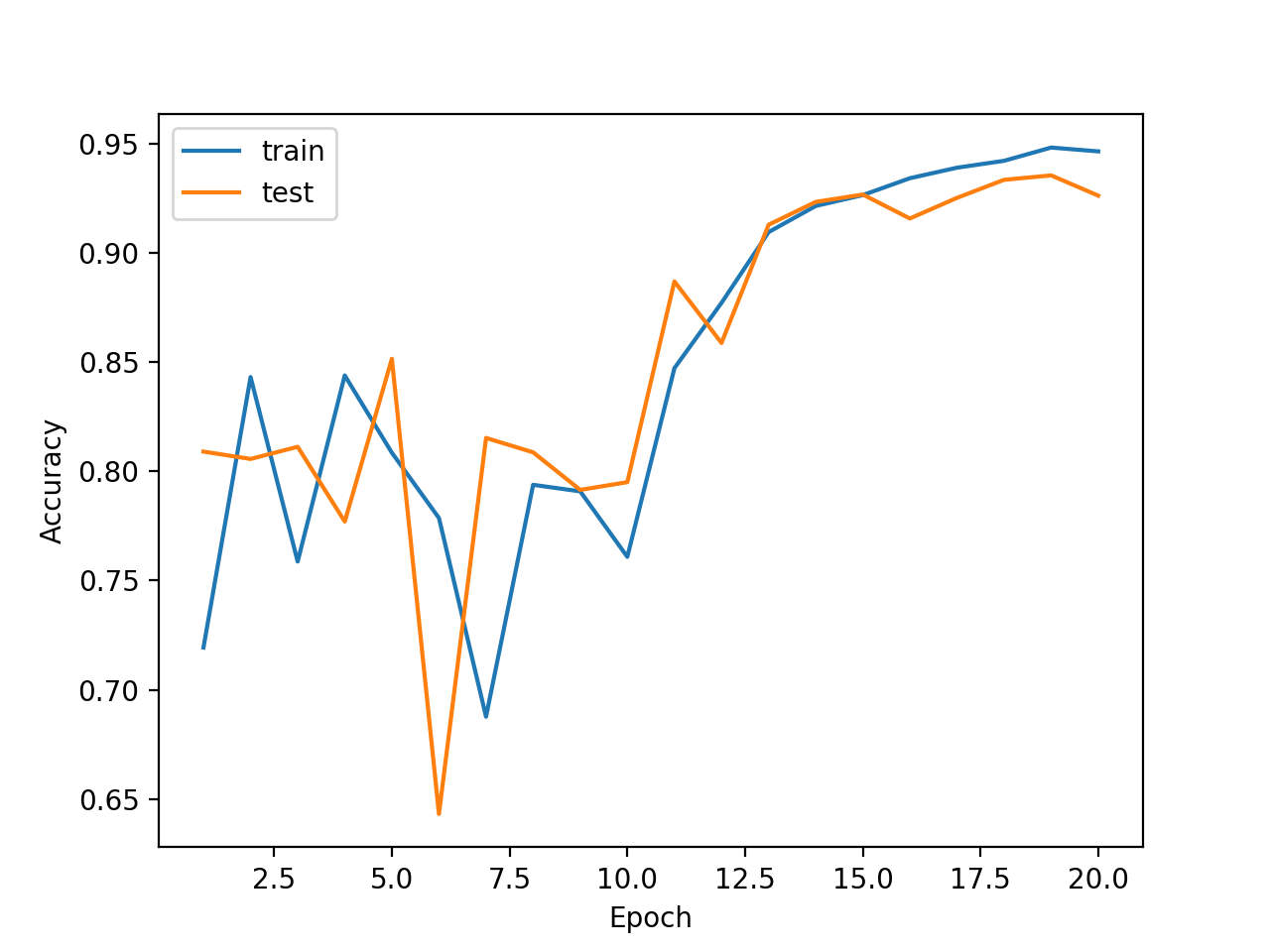}\\

The results seem solid, but this approach is not included
in our model because it takes too much time to run and we
do not have time to fine-tune the hyperparameters. But in future work, we believe that using LSTMs with real recurrence will give an even better results.

While we achieve great performance in this dataset, the question remains as to whether X (to be replaced by the best model) can still perform well in tasks that classify news into more than two categories, such as the Fake News Challenge. In that case, a simple unidirectional LSTMs may not be so well and may need to be replaced by a bidirectional one. In addition, it would be interested to know how well our pre-trained model performs in other downstream tasks, such as Spam Detection. Lastly, in our model, the pre-training is done on the dataset given (will make the model specific to the task), instead of on the big corpus available online, such as Google's pre-trained Word2Vec model. If the task were a classification of four or eight categories, pre-trained model on large corpus may perform better as the model is pre-trained on more words.

We can also try to improve the training by using different word embeddings. While we only chose only 3 different types of embeddings, we could have tried different embeddings such as GloVe and the features used are entirely dependent only on context words. We can use different forms for encoding texts which can be used to be trained using these algorithms to achieve a better model. In another

State-of-the-art pre-trained models can be used if the task is no longer a binary classification. Models like Transformer and BERT will be strong candidates as they have learned a very strong representation that takes the context into account when computing an embedding for a word. Unlike LSTMs whose sequential nature prohibits parallelization, the Transformer and the BERT can achieve parallelization by replacing recurrence with the attention mechanism. Thus, they require less computation power and can be easily fine-tuned in downstream tasks.

\section{Appendix}

\section*{Github Repo}
\url{https://github.com/Sairamvinay/Fake-News-Dataset}

\section*{Author Contributions}

Sairamvinay Vijayaraghavan: Project Planning, Problem Formation, DataSet Search, POS Distribution graph, Code for CountVectorizer, Word2Vec, ANN, Randomforest,To parse csv files (readdata), Code integration for TextVectorizer, Grid Search model running, ROC model running, Code Base Cleanup and management (further cleanup), PowerPoint Checking, Report Analysis for W2V, ANN, Report editing

Zhiyuan Guo: Project Planning, DataSet Search, Polarity Graphs, Code for LSTM, RandomForest, Adding Functionality and Readability in each of the scripts, Code Integration, Grid Search model running, ROC model running, PowerPoint Development, Report Analysis for TFIDF and LSTM, Report Analysis for the Abstract, the Discussion, Conclusion, Pipeline Diagram, Report editing

Ye Wang: Project Planning, DataSet Search, Code for TFIDF, PCA, Grid Search model running, ROC model running, Report Integration into Latex, Report Analysis of the Results (table creations), Report Analysis for the Outlier Removal, Random Forest, Report editing

John Voong: Word2Vec, DataCleanup (StopWord Cleanup), Grid Search model running, ROC model running, PowerPoint Development, Report Analysis for W2V, Pipeline Diagram, Report editing, Paper structure

Wenda Xu: Code for PCA, ROC model running, Code Base Cleanup and management, PowerPoint Development, Report Analysis about Count Vectorizer, Report Analysis about Logistic Regression

Armand Nasseri: Project Planning, Dataset search, Code for SVM, Data Cleanup (StopWord Cleanup), ROC model running, PowerPoint Development, Report Analysis about SVM

Jiaru Cai: Outlier Removal, Accuracy and Loss Plots for Neural Network, PowerPoint Framework

Kevin Vuong: DataCleanup (remove punctuations), Code for Logistic Regression, Grid Search model running, PowerPoint Cleanup, Report Analysis about Data Cleanup, Introduction and Abstract

Linda Li: Unigram and Bigram analysis, Code for ROC plots, Report Analysis of the Data Cleanup section, Graph analysis

Eshan Wadhwa: Related Work, References and Citation (Introduction and Field research), Report Editing, PowerPoint slides,

\section*{References}
\small

[1] Samir Bajaj, “The Pope Has a New Baby!” Fake News Detection Using Deep Learning”, Winter 2017,\\ \url{https://pdfs.semanticscholar.org/19ed/b6aa318d70cd727b3cdb006a782556ba657a.pdf}

[2] Arjun Roy, Kingshuk Basak, Asif Ekbal, and Pushpak Bhattacharyya, “A Deep Ensemble Framework for Fake News Detection and Classification”, 12 November 2018, \\ \url{https://arxiv.org/pdf/1811.04670.pdf}

[3] Niall J. Conroy, Victoria L. Rubin, and Yimin Chen, “Automatic Deception Detection: Methods for Finding Fake News”, November 2015,\\ \url{https://asistdl.onlinelibrary.wiley.com/doi/epdf/10.1002/pra2.2015.145052010082}.

[4] Liang Wu and Huan Liu, “Tracing Fake-News Footprints: Characterizing Social Media Messages by How They Propagate”, February 2018,\\ \url{http://www.public.asu.edu/~liangwu1/WSDM18_TraceMiner.pdf}

[5] Adrian Colyer, “Tracing fake news footprints: characterizing social media messages by how they propagate”,the morning paper, February 2018, 
\url{https://blog.acolyer.org/2018/02/19/tracing-fake-news-footprints-characterizing-social-media-messages-by-how-they-propagate/}

[6] Kai Shu, Amy Sliva, Suhang Wang, Jiliang Tang and Huan Liu, “Fake News Detection on Social Media: A Data Mining Perspective”, August 2017,\\ \url{https://arxiv.org/abs/1708.01967}

[7] Jiawei Zhang, Bowen Dong and Philip S. Yu, “FAKEDETECTOR: Effective Fake News Detection with Deep Diffusive Neural Network”, August 2019,\\ \url{https://arxiv.org/pdf/1805.08751.pdf}

[8] Sepp Hochreiter and Jurgen Schmidhuber, “Long short-term memory”, November 1997,\\
\url{http://www.bioinf.jku.at/publications/older/2604.pdf}

[9] Yoshua Bengio, Patrice Simard, and Paolo Frasconi. “Learning long-term dependencies with gradient descent is difficult”, March 1994,\\ \url{http://www.comp.hkbu.edu.hk/~markus/teaching/comp7650/tnn-94-gradient.pdf}

[10] Gaofeng Cheng, Vijayaditya Peddinti, Daniel Povey, et al., “An Exploration of Dropout with LSTMs”. August 2017,\\
\url{https://www.danielpovey.com/files/2017_interspeech_dropout.pdf}

[11] Juan Ramos. “Using tf-idf to determine word relevance in document queries”, December 2003,\\
\url{https://www.cs.rutgers.edu/~mlittman/courses/ml03/iCML03/papers/ramos.pdf}

[12] Gerard Salton and Christopher Buckley. “Term-weighting approaches in automatic text retrieval”, January 1988,\\ \url{https://www.sciencedirect.com/science/article/abs/pii/0306457388900210}

[13] Jason Brownlee. “How to Prepare Text Data for Machine Learning with scikit-learn”, August 2019, \\
\url{https://machinelearningmastery.com/prepare-text-data-machine-learning-scikit-learn/}
\end{document}